\documentclass[preprint]{elsarticle}

\usepackage{graphicx}
\usepackage{amssymb}

\usepackage{lineno,hyperref}
\usepackage{amsmath}
\usepackage{subcaption}
\usepackage{natbib}
\usepackage{commath}
\usepackage{tabularx, booktabs}
\usepackage{multirow}
\usepackage[para,online,flushleft]{threeparttable}
\usepackage{rotating}
\usepackage{color,soul}

\usepackage{url}
\usepackage[para,online,flushleft]{threeparttable}

\modulolinenumbers[5]

\journal{Journal of Pattern Recognition Society}

\begin{document}

\begin{frontmatter}
\setlength{\abovedisplayskip}{0pt}
\setlength{\belowdisplayskip}{0pt}
\setlength{\abovedisplayshortskip}{0pt}
\setlength{\belowdisplayshortskip}{0pt}
\title{A Recursive Bayesian Approach To Describe Retinal Vasculature Geometry}

\author[mymainaddress]{Fatmat\"{u}lzehra Uslu\corref{mycorrespondingauthor}}
\cortext[mycorrespondingauthor]{Corresponding author}
\address[mymainaddress]{Bioengineering Department, Imperial College London,South Kensington Campus, SW7 2AZ, UK}
\ead{fzehrauslu@gmail.com}

\author[mymainaddress]{Anil Anthony Bharath}
\ead{a.bharath@imperial.ac.uk}

\begin{abstract}
Demographic studies suggest that changes in the retinal vasculature geometry, especially in vessel width, are associated with the incidence or progression of eye-related or systemic diseases. To date, the main information source for width estimation from fundus images has been the intensity profile between vessel edges. However, there are many factors affecting the intensity profile: pathologies, the central light reflex and local illumination levels, to name a few. In this study, we introduce three information sources for width estimation. These are the probability profiles of vessel interior, centreline and edge locations generated by a deep network. The probability profiles provide direct access to vessel geometry and are used in the likelihood calculation for a Bayesian method, particle filtering. We also introduce a geometric model which can handle non-ideal conditions of the probability profiles. Our experiments conducted on the REVIEW dataset yielded consistent estimates of vessel width, even in cases when one of the vessel edges is difficult to identify. Moreover, our results suggest that the method is better than human observers at locating edges of low contrast vessels.
\end{abstract}

\begin{keyword}
Particle filtering \sep Deep Belief Nets \sep deep neural networks  \sep fundus images \sep width estimation \sep tracking
\end{keyword}

\end{frontmatter}


\section{Introduction}
The retina provides a convenient way to image fine vasculature optically. Both organ-specific diseases, such as diabetic retinopathy or glaucoma and also systemic disorders such as diabetes, hypertension and cardiovascular diseases induce early changes in the retina. Moreover, the brain, which is closely located to the eye, shares similar characteristics to retinal vasculature. 

Potential associations between the changes on the retinal vasculature geometry and the presence of some diseases have been the subjects of demographic studies \cite{ikram2013retinal,ding2014retinal,gopinath2014associations,cheung2014microvascular}. The changes on the vasculature, which may be subtle, can be on vessel width, curvature and branching angles. In order to detect the changes on the vasculature, one needs to analyze it quantitatively. This involves steps beyond mere segmentation: such as estimating the locations of vessel centerlines and edges to sub-pixel precision, at a basic level, and vessel widths and curvatures, at a more sophisticated level.  However, the quality of fundus images captured in large screening programmes varies. The variable image quality can present a challenge even to the best segmentation algorithms, let alone any attempt to obtain automatic estimates of vessel width. This may be sometimes complicated by the complex topology and global geometry of the vasculature, where vessels can be very close or overlap; moreover, vessel appearance may be affected by pathologies, the central light reflex or uneven illumination.

In the literature, the majority of methods for the quantitative analysis of vasculature use binary vessel maps obtained by segmentation as {\it a priori} information to locate vasculature. Techniques for obtaining quantifiable measures include model fitting \cite{lowell2004measurement,lupacscu2013accurate, araujoa2017estimation,aliahmad2016adaptive,gao2001method}, graphs \cite{xu2011vessel} or active contours \cite{al2009active}. The location of pixels inside vessels can be obtained from vasculature skeletons generated by thinning binary vessel maps \cite{lupacscu2013accurate,araujoa2017estimation}. Because of reliance on segmentation, these approaches may not be aware of missing or false vessels inherited from the segmentation.

On the other hand, a few methods, not included in the former group, require prior knowledge of vessel parameters only to start a largely autonomous estimation process. For example, tracking methods sequentially estimate the vessel trajectory and geometry parameters, given a prior estimate of the vessel location and parameters \cite{chutatape1998retinal,yin2012retinal,zhang2014retinal,wu2016deep}. These approaches are well-aligned with the principles of Bayesian estimation. As an example, Chutatape \textit{et al.} \cite{chutatape1998retinal} used an extended Kalman filter, Yin \textit{et al.} and Zhang \textit{et al.} Maximum a Posteriori (MAP) and, Wu \textit{et al.} generalized particle filters \cite{wu2016deep}. In all of these approaches, a key component influencing the performance of tracking seems to relate to how well the likelihood function reflects the actual vessel geometry and appearance in the original image data \cite{yin2012retinal,zhang2014retinal}.

To date, approaches to Bayesian vessel tracking have used likelihood functions that describe the appearance of vessel cross-section in the original, or at most band-pass filtered, image data. For example, the cross-sectional vessel intensity profile has been approximated with one-dimensional Gaussian functions, \cite{chutatape1998retinal,yin2012retinal,zhang2014retinal}. However, the intensity profile can be easily affected by many factors, such as the presence of pathologies, the central light reflex, uneven illumination of the retina, noise, the contrast of vessels, the focus of the camera; the latter may lead choroidal vessels being superimposed on retinal vasculature. These factors may make the Gaussian approximation too optimistic for vessel parameter estimation. As reported by Zhang \textit{et al.} \cite{zhang2014retinal}, adding vesselness information in longitudinal direction to the likelihood function led to an improvement in the tracking performance obtained by Yin \textit{et al.} \cite{yin2012retinal}, who used only vessel profiles in the likelihood function.

In contrast, this study introduces a new way to represent vessel cross-sections: using probability maps for  vessel interior, centerline and edge locations. The probability maps are produced by a single deep network, thus maintaining the relations between the vessel parts, significantly simplifying the construction of a likelihood function. Then, the likelihood function is used in a sequential Bayesian method, particle filtering, which extracts a comprehensive and precise representation of vasculature, through the estimation of parameters that describe vessel geometry. We introduce a new model for particle filtering to improve the flexibility in the search of best fitting parameters to actual vessel geometry. 

\begin{figure}[!ht]
\centering
 {\includegraphics[scale= 0.15,trim=0cm 0cm 0cm 0cm,clip,keepaspectratio]{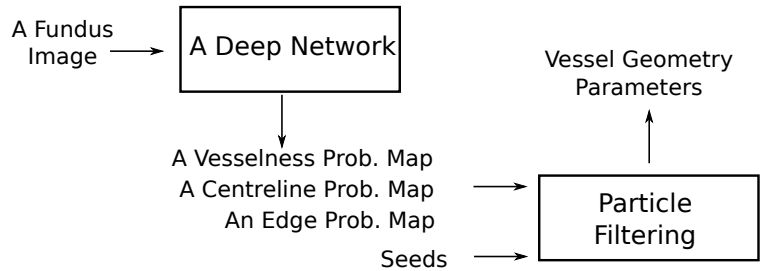}}
  \caption{A general overview of the proposed method. The particle filtering operates directly on the probability maps output by a deep network.  \label{fig:GeneralOverview}}
\end{figure}

Finally, we suggest a new way to evaluate the performance of tracking in vasculature, which considers the dependency of vessel profiles in a vessel segment. This approach to performance evaluation is suggested to better identify vessels where there is large disagreement between reference and estimated values. The experiments indicate that the method can cope with various characteristics of a dataset without re-training the network on unseen datasets. Moreover, where vessels are very small and of low contrast, the method appears to be more reliable than human observers at detecting vessel boundaries. Figure \ref{fig:GeneralOverview} presents a general overview of the approach.  

\section{A Probabilistic Tracking Method For Retinal Vasculature}
\subsection{Problem Definition}
Tracking of retinal vasculature can be described as a recursive estimate of vessel geometry parameters, considering the smooth variations on vessel thickness and curvature over a vessel branch. The recursion can be initiated with an initial estimate of the geometry parameters given at iteration $k=\mathtt{0}$.  For example, at $k=2$, the geometry parameters are initially estimated given the parameter estimates at $k=1$, and then corrected by evaluating the fitness of the initial predictions to measurements obtained from the vessel part under tracking.  The evolution of the geometric parameters and the relation between the geometry parameters and the measurements can be modeled with (\ref{eqn:StateUpdate1}-\ref{eqn:StateUpdate2}). 

\begin{align} 
&\mathbf{z}_{k+1}=G(\mathbf{z}_{k}, \mathbf{v}_{k})\label{eqn:StateUpdate1}\\
&\mathbf{y}_{k+1}=O(\mathbf{z}_{k+1}, \mathbf{u}_{k+1}) \label{eqn:StateUpdate2}
\end{align}
 
where $G(\cdot)$ is a model of vessel geometry, which captures the evolution during tracking of the parameters describing vessel geometry in a recursive way. $O(\cdot)$ is an observation model, which relates a set of the geometry parameters to a set of measurements. The state vector $\mathbf{z}$, contains geometric parameters describing the vessel being tracked, and $\mathbf{v}$ and $\mathbf{u}$, respectively, represent the uncertainty in the geometry and observation models.  

\subsection{Bayesian Approach to Solution}
The estimation of the geometry parameters given the observations can be defined with the prediction of the posterior probability distribution of the geometry parameters $P(\mathbf{z}_{k+1}|y_{k+1})$ from the Bayesian point of the view. The posterior probability distribution can be calculated from Bayes rule in (\ref{eqn:BayesRule}). 

\begin{equation} \label{eqn:BayesRule}
P(\mathbf{z}_{k+1}|\mathbf{y}_{0:k+1}) \propto P(\mathbf{z}_{k+1}|y_{0:k}) \cdot P(\mathbf{y}_{k+1}|\mathbf{z}_{k+1})
\end{equation}
where $P(\mathbf{z}_{k+1}|y_{0:k})$ and $P(\mathbf{y}_{k+1}|\mathbf{z}_{k+1})$ respectively show the prior probability distribution of the geometry parameters and their likelihood at iteration $k+1$. 

The prior probability distribution reflects our belief about vessel parameters, which is represented with the updated initial estimate of the geometry parameters at the start of each iteration, $k$. When $k=0$, the prior probability distribution could be initialised by manual input, or a method that detects vessel tracks as they leave the optic disc \cite{zhang2014retinal}. As iterations proceed, the prior probability distribution could be evolved according to the geometry model $G(\cdot)$ in (\ref{eqn:StateUpdate1}), where the posterior probability distribution of the geometry parameters at iteration $k$ is used as the prior probability distribution of the geometry parameters for iteration $k+1$. The role of the noise, $\mathbf{v}$, in (\ref{eqn:StateUpdate1}) is to explain how much change in vessel geometry is foreseen over an iteration. 

The prior probability distribution over geometric parameters could then updated with (\ref{eqn:BayesRule}), according to the fitness of this distribution to measurements obtained from the image, described by the likelihood. The likelihood, $P(\mathbf{y}_{k+1}|\mathbf{z}_{k+1})$ incorporates the observation model, $O(\cdot)$, in (\ref{eqn:StateUpdate2}).

\subsubsection{Particle Filtering}
Particle filtering, a technique based on recursive Bayesian estimation, can capture an arbitrary posterior probability distribution of geometry parameters at iteration $k+1$, with a set of particles $ \{\mathbf{z}^{n}_{k+1}\}_{n=1}^{N}$ and their weights $ \{W^{n}_{k+1}\}_{n=1}^{N}$. Each particle, $n$, hypothesizes a set of the geometry parameters at iteration $k+1$ $\mathbf{z}^{n}_{k+1}$. According to the Law of Large Numbers \cite{grinstead2012introduction}, a sufficient number of particles can approximate the distribution; $P(\mathbf{z}_{k+1}|\mathbf{y}_{k+1}) \approx \sum_{n=1}^{N} W^{n}_{k+1} \delta (\mathbf{z}-\mathbf{z}^{n}_{k+1})$.

Particle filtering frequently uses importance sampling \cite{arulampalam2002tutorial} to estimate the posterior probability distribution. Importance sampling initially samples particles from a proposal distribution, then the weights of the particles are updated with the importance weights \cite{doucet2001introduction}. In the case of the proposal distribution being the prior probability distribution, and using resampling after each weight update, the recursive updates to the importance weights is simplified to (\ref{eqn:WeightsLikelihood2}) \cite{arulampalam2002tutorial}:  

\begin{equation} \label{eqn:WeightsLikelihood2}
W_{k+1}^{n}  \propto P(y_{k+1}|\mathbf{z}_{k+1}^{n}) 
\end{equation}
where $W_{k}^{n}$ and $W_{k+1}^{n}$ are the weights of $n^{th}$ particle at iteration $k$ and $k+1$ consecutively. $P(y_{k+1}|\mathbf{z}_{k+1}^{n})$ denotes the likelihood of the $n$-th particle at iteration $k+1$.

The expectation of the posterior probability distribution can be calculated with (\ref{eqn:Expectation}), returned as the geometry parameters estimated at iteration $k+1$: 
\begin{equation} \label{eqn:Expectation}
\mathbf{\bar{z}} \approx \sum_{n=1}^{N} W^{n}_{k+1} \mathbf{z}^{n}_{k+1}
\end{equation}

\subsection{A Geometry Model}
The three-dimensional shape of a vessel can be assumed to be a tube with a width (diameter) $w$ and a centreline location $C$, and oriented in a direction $D$. The change on its geometry over a small distance $s$ (related to the step size during tracking) can be assumed to be smooth, even though sudden changes can rarely occur due to pathologies such as vessel beading. Regarding the geometry model in (\ref{eqn:StateUpdate1}), the change on the geometry can be modelled by using a normal distribution, which accounts for the noise $\mathbf{v}$. Although we can model the vessel geometry with the diameter $w$, centreline location $C$, direction $D$ and step size $s$ with the tube model, it may be difficult to infer these parameters from the appearance of vessels in fundus images without considering factors influencing its appearance such as imaging, noise and pathologies.

To date, many Bayesian tracking methods \cite{chutatape1998retinal,yin2012retinal, zhang2014retinal} have modelled the appearance of the cross-section of a vessel segment conditional on centerline location, diameter and orientation, by Gaussian functions. Though analytically convenient, it is not realistic when the shape of the intensity profile changes due to uneven illumination, pathologies or other noise components. The differences between the shape of the intensity profile of a vessel without the central light reflex, and a Gaussian function describing it may appear as (i) the intensity profile is skewed to one of vessel edges \cite{lupacscu2013accurate} and (ii) the intensities at edges of the same vessel may be different \cite{araujoa2017estimation}.  These characteristics of the intensity profile have been addressed by Lupacscu \textit{et al.} and Araujoa \textit{et al.}\cite{lupacscu2013accurate,araujoa2017estimation} in their parametric models to improve the estimates of vessel widths. However, these characteristics have not been applied in Bayesian tracking methods. 

In this study, we factorise the probability profile of vessel cross-section to vessel centreline and edge probability profiles. This factorisation allows us to relax the symmetry in appearance models implied in using Gaussians to model vessel profile; in other words, the peaks of Gaussian functions being in the middle of the intensity profile. Therefore, we can model a skewed vessel appearance over a profile by using two parameters for vessel width: the distance between an arbitrary location, $A$, inside the vessel and left edge $wl$ and that between the location and right edge $wr$. The state vector of the vessel parameters at iteration $k$ can be written $\mathbf{z}_{k}=[z^{d}_{k}, z^{A}_{k}, z^{wl}_{k},z^{wr}_{k}]$. The evolution of the state vector over iterations is given with (\ref{eqn:MotionModel3})- (\ref{eqn:MotionModel5_2}):

\begin{align}
&z^{D}_{k+1}=z^{D}_{k}+\epsilon_{z^{D}} \label{eqn:MotionModel3}\\
& z^{A}_{k+1}= z^{A}_{k} +z^{D}_{k+1} \cdot s +\epsilon_{z^{A}} \label{eqn:MotionModel1}\\
&z^{wr}_{k+1}=z^{wr}_{k}+\epsilon_{z^{wr}} \label{eqn:MotionModel5_1}\\
&z^{wl}_{k+1}=z^{wl}_{k}+\epsilon_{z^{wl}} \label{eqn:MotionModel5_2} 
\end{align} 
where $\epsilon_{z}$ represents a normal distributed noise variable ($\epsilon_{z^{D}}$ for direction vector, $\epsilon_{z^{wr}}$ and $\epsilon_{z^{wl}}$ for width and $\epsilon_{z^{A}}$ for the arbitrary interior location).  $s$ is a constant and denotes the step size for tracking.  $(\mathbf{x})_{\bot} $ denotes a direction vector perpendicular to  arbitrary vector, $\mathbf{x}$.

\subsection{An Observation Model}

\begin{figure}[!ht]
   \centering
   {\includegraphics[scale= 0.65,trim=0cm 0cm 0cm 0cm,clip,keepaspectratio]{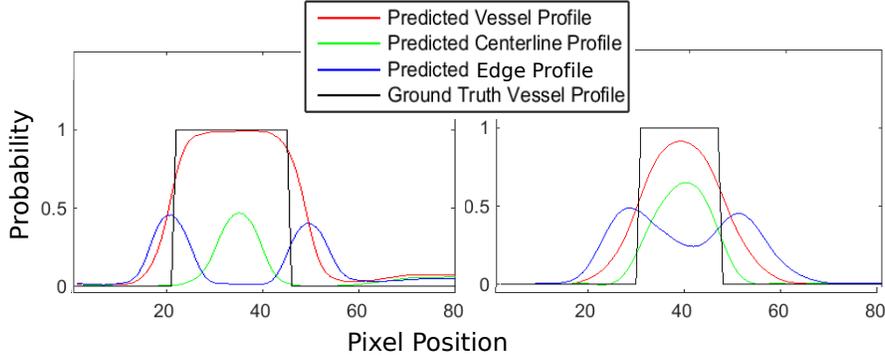}}
  \caption{Probability profiles in the left side belong to the larger vessel and those in the right side to the thinner one in Figure \ref{fig:LabelPatches}. (Best viewed in color.) \label{fig:VesselProfileComparision} }
\end{figure}

In this study, we introduce three sources of information, namely: from vessel centreline, edge, and interior probability maps. How these maps are generated will be explained in Section \ref{sec:GenProbMap}. We utilise profiles through the edge and centreline probability maps, rather than from the intensity images, to perform tracking. Figure \ref{fig:VesselProfileComparision} compares the probability profiles for a large and thin vessel, obtained along the red lines from the  centreline and edge probability maps shown in Figure \ref{fig:LabelPatches}, with those taken from reference vessel maps. The former figure exemplifies the typical characteristics of the probability profiles: (i) these curves are virtually free of either noise components, or intensity variations that are present within the fundus images in Figure \ref{fig:LabelPatches}. (ii) these curves have maxima at locations which almost overlap with significant points (e.g. centerline locations) of the reference vessel profiles. The small disagreement between reference and estimated profiles is acceptable: locating precise boundaries is difficult even for human observers, and subject to significant inter-observer variability \cite{staal2004ridge,hoover2000locating,al2009active}.

\begin{figure}[!ht]
\centering
 {\includegraphics[scale= 0.6,trim=0cm 0cm 0cm 0cm,clip,keepaspectratio]{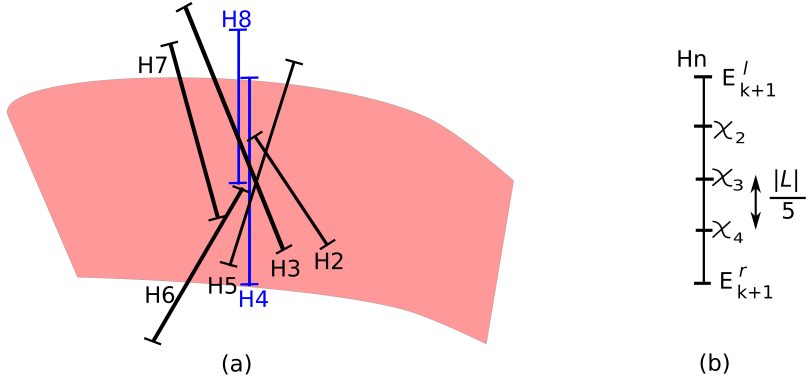}}
\caption{(a) Some hypotheses over a vessel part sampled from the prior probability distribution according to (\ref{eqn:MotionModel3})-(\ref{eqn:MotionModel5_2}). Each search line accounts for a set of hypothesized vessel parameters (e.g. centreline location, width and direction) and is named with $H$. In order to show hypothetic vessel edges represented by each hypothesis, the lines are accompanied with perpendicular short lines at hypothetic edges. (b) Edges ${E}^{l,r}_{k+1}$ and the sampling locations $\chi_{j}$; $j=1:3$ in (\ref{eqn:LikelihoodModel1}) are shown on the line $Hn$. The distance between neighbouring sample locations given a particular hypothesis is the same and equal to one fifth of the length of the line $|\mathit{L}|$. (Figure best viewed in color.)\label{fig:Hypotheses}}
   \end{figure}

The probability profiles are sampled from an arbitrary search region. Apart from Yin \textit{et al.} \cite{yin2012retinal} who used an adaptable semi-elliptical search curve for each iteration, our search region contains hundreds of adaptable lines (see Figure \ref{fig:Hypotheses}(a)). The spatial distribution of these lines is driven by the normally distributed noise in (\ref{eqn:MotionModel3})-(\ref{eqn:MotionModel5_2}). The ends of a line correspond to hypothesised edge locations, calculated with (\ref{eqn:EdgeCalculation0100}) with respect to a parameter set sampled from the prior probability distribution. A search line is divided into $4$ segments of equal length by sampling at $3$ predetermined locations. The locations were selected in a way that they can capture the overall shape of the probability profiles illustrated in Figure \ref{fig:VesselProfileComparision}.

The observation model, providing the likelihood of each hypothesis, is given in (\ref{eqn:LikelihoodModel1}). 

\begin{align} \label{eqn:LikelihoodModel1}
&P(y_{k+1}|\mathbf{z}_{k+1})= P_{e}^{'}\cdot P_{c}^{'} \cdot  P_{c}(z^{A}_{k+1})  \cdot P_{s}\\
&P_{e}^{'}= \prod_{i=\{l,r\}} P_{e}({E}^{i}_{k+1}) \cdot (1-P_{c}({E}^{i}_{k+1})) \nonumber\\
&P_{c}^{'}= \prod_{j=\{1,2,3\}} P_{c}(\chi=\chi_{j}) \cdot (1-P_{e}(\chi=\chi_{j}))  \nonumber\\
&P_{s}=\abs{z^{D}_{k+1} \cdot \overrightarrow{E}(z^{A}_{k+1}) } \nonumber
\end{align} 
where $P_{e}$ and $P_{c}$ are profiles respectively obtained from edge and centerline probability maps between hypothesized edge locations ${E}^{l}_{k+1}$ and ${E}^{r}_{k+1}$. $P_{s}$ denotes the similarity of hypothesized vessel direction $z^{D}_{k+1}$ to eigenvector $\overrightarrow{E}(z^{A}_{k+1})$ at hypothesized arbitrary interior location $z^{A}_{k+1}$. Sampling locations from the probability profiles are 
$\chi_{1}=\frac{({3E}^{l}_{k+1}+{E}^{r}_{k+1})}{4}$, $\chi_{2}= \frac{({E}^{l}_{k+1}+{E}^{r}_{k+1})}{2}$ and $\chi_{3}= \frac{({E}^{l}_{k+1}+{3E}^{r}_{k+1})}{4}$ 
(see Figure \ref{fig:Hypotheses}(b)).

Equation (\ref{eqn:LikelihoodModel1}) has three main components to be maximized: $P_{e}^{'}$, $P_{c}^{'}$ and $P_{s}$. The first one calculates the probability of edge locations and increases when the edge estimates of the tracker become closer to vessel edge locations, where edge probabilities and the complements of centerline probabilities are the maximum. The second one aims to maximize centerline probability in order to make sure of the centreline estimates of the tracker to be inside the vessel under tracking. This component is important in terms of avoiding the tracker to trace boundaries of two different vessels in a close distant. When the centreline estimates of the tracker is inside the vessel, $P_{c}^{'}$ is far larger than when it is between vessels. This is due to the centerline probabilities and the complement of edge probabilities to be much larger inside the vessel and almost zero outside the vessel. The third component measures the similarity between an eigenvector indicating the direction of the vessel and the direction hypothesized by the tracker. Both vectors are obtained from the vessel interior probability map at hypothesized centerline locations. Even if the hypothesized edge locations fall on the edges of a vessel, this does not guarantee that the tracker will estimate accurate geometry parameters unless the orientation of the probability profiles consistently aligns with the orientation of the vessel cross-section.  Therefore, the term $P_s$ in (\ref{eqn:LikelihoodModel1}) contributes to the consistency of width estimations by assigning larger likelihoods to hypothesized edge locations that have similar alignment to the vessel cross-section. 
\subsubsection{Estimating Vessel Edges}
After calculating the expectation of the posterior probability distribution with (\ref{eqn:Expectation}), the locations ${E}_{k+1}^{l,r}$ for the left and right edges of a vessel may then be calculated using the state parameters according to
\begin{equation} \label{eqn:EdgeCalculation0100}
{E}_{k+1}^{l,r}= z^{A}_{k+1}\pm z^{wl,wr}_{k+1} \cdot (z^{D}_{k+1})_{\bot}  
\end{equation}

\subsubsection{Strong and Weak Hypotheses}
\begin{figure}[!ht]
 {\includegraphics[scale= 0.4,trim=0cm 0cm 0cm 0cm,clip,keepaspectratio]{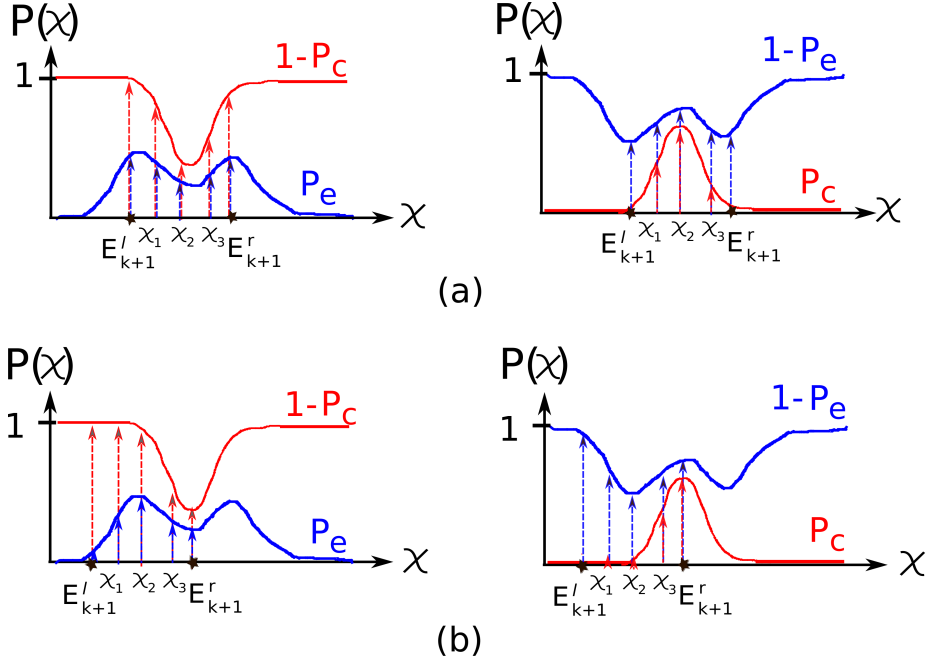}}
\caption{The calculation of likelihoods of two hypotheses, on (a) for a strong and (b) for a weak hypothesis, which are shown with $H4$ and $H8$ in Figure \ref{fig:Hypotheses}(a) consecutively.  $E_{k+1}^{l}$ and $E_{k+1}^{r}$ denote hypothesized edge locations and $\chi_{1}$, $\chi_{2}$ and $\chi_{3}$ sampling locations in (\ref{eqn:LikelihoodModel1}). Centerline probability profile $P(c)$ and its complement $1-P(c)$ are shown with red solid lines, and samples from these profiles with red arrows. Similarly, an edge profile $P(e)$ and its complement $1-P(e)$ are demonstrated with blue solid lines, and samples with blue arrows. The right and left plots respectively represent the calculation of $P_{e}^{'}$ and $P_{c}^{'}$ in (\ref{eqn:LikelihoodModel1}). Ignoring the effect of $P_{c}(z^{A}_{k+1}) \cdot P_{s}$ in (\ref{eqn:LikelihoodModel1}), the likelihood of the strong hypothesis is a million times larger than that of the weak hypothesis. (Figure best viewed in color.) \label{fig:SampledProfile} }
\end{figure}

Figure \ref{fig:SampledProfile} exemplifies the likelihoods of one strong and one weak hypothesis. The strong hypothesis estimates edge locations closer to the peaks of the edge probability profile, while the weak one predicts one of the edges outside the vessel and the other inside the vessel. In order to simplify the comparison of their likelihoods, two assumptions are made for both hypotheses:  (i) the orientation of hypothesized vessel cross-sections is the same, (ii) hypothesized centerline locations are in the middle of the hypothesized edge locations. Considering the heights of the arrows denoting probabilities, it is obvious that the tracker assigns significantly far lower likelihood to the weak hypothesis (Figure \ref{fig:SampledProfile} (b)) than the strong hypothesis (Figure \ref{fig:SampledProfile} (a)) by reducing its contribution to the expectation of the posterior probability distribution; it is this weighting that ultimately determines the eventual estimate of the geometry parameters.  

\begin{figure}[!ht]
  \begin{tabular}{c c }
  \begin{subfigure}[b]{0.5\textwidth}
  \centering
 {\includegraphics[scale=0.25,
]{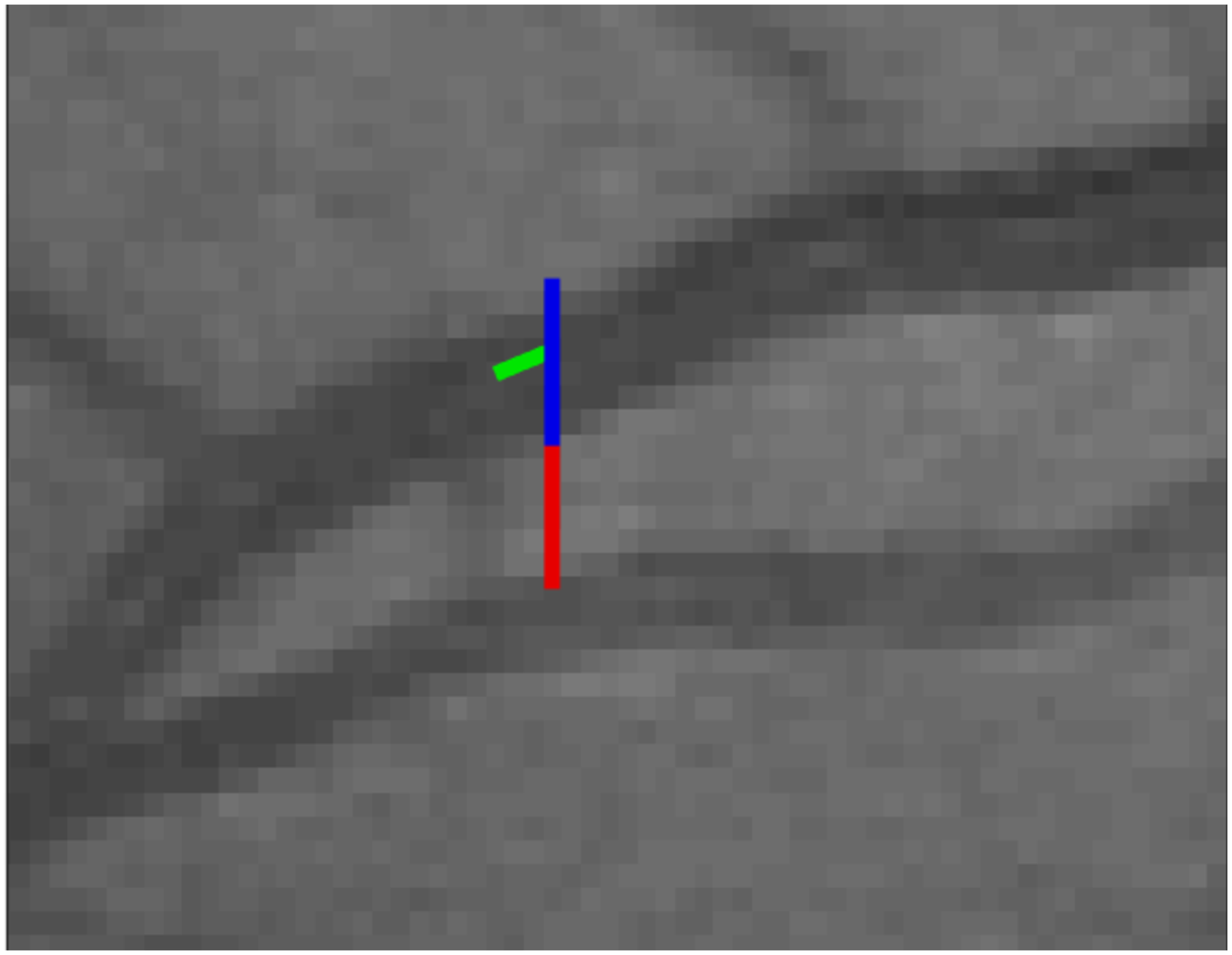}}
  \caption{}
 \end{subfigure}
 &
  \begin{subfigure}[b]{0.5\textwidth}
  \centering
 {\includegraphics[scale=0.3]{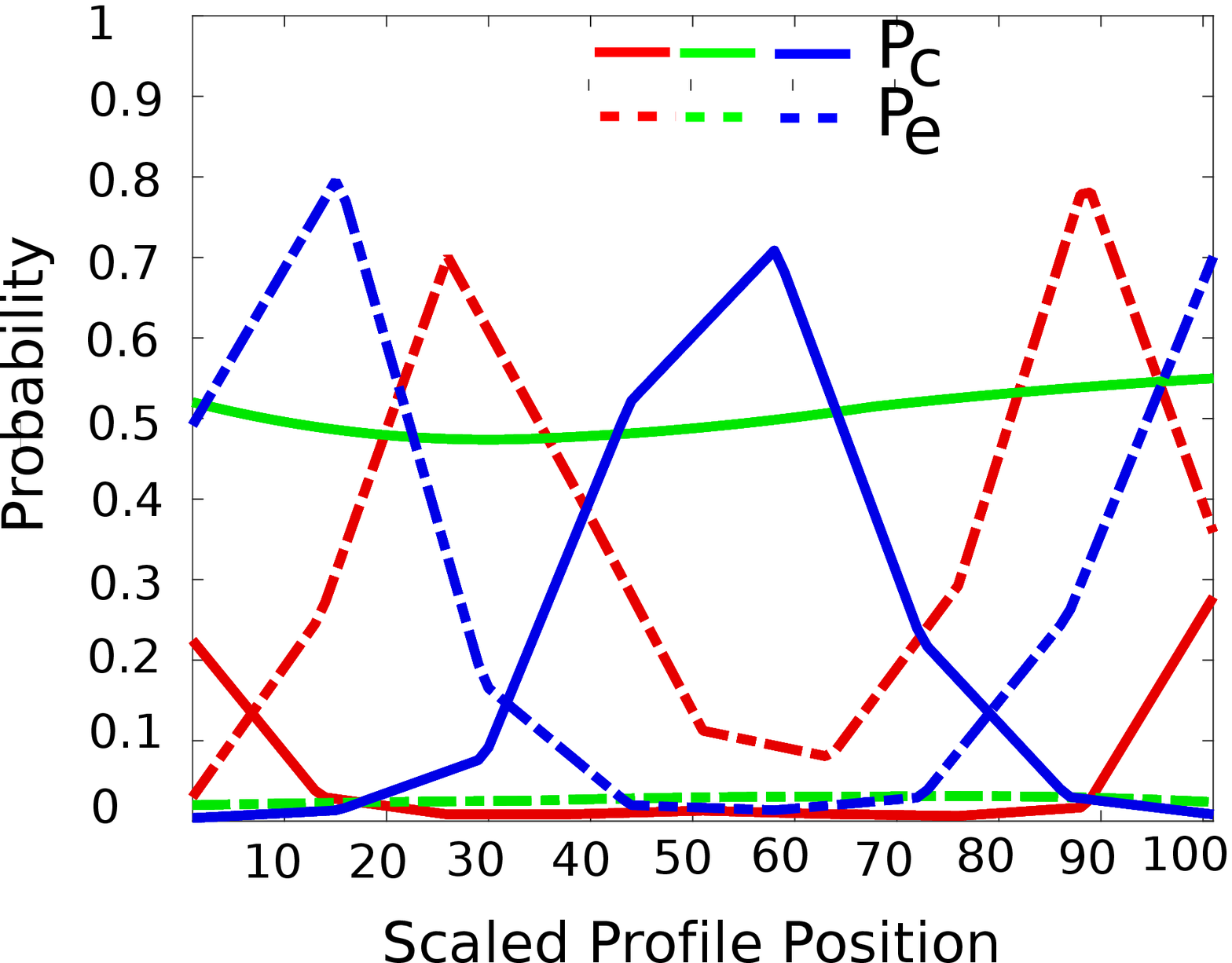}}
\caption{}
 \end{subfigure}
 \end{tabular}
  \caption{Search lines and their likelihoods:(i) the blue search line is between edges of the vessel under tracking, (ii) the red search line is between edges of two vessels and (iii) the green search line is inside the vessel but oriented parallel to it. (b) Corresponding edge $P_{e}$ and centerline $P_{c}$ probability profiles are indicated by dashed and solid lines consecutively, and with colors matching those of (a). The likelihood is $1.00e^{-5}$ for the blue one, $9.15e^{-12}$ for the red one and $9.56e^{-8}$ for the green one.   
Note that $P_{s}$ in (\ref{eqn:LikelihoodModel1}) was not included in the likelihood calculation of these lines. (Figure best viewed in color.) \label{fig:WrongEstimations}}
\end{figure}

Figure \ref{fig:WrongEstimations} illustrates three cases, where (i) the search line in blue finds relatively better edge locations, (ii) the search line in red is located between edges of different vessels and (iii) the search line in green is oriented parallel to the vessel. Among these search lines, the blue one has significantly larger likelihood. Also, the likelihood of the red line is much lower than that of the green line, which indicates that the proposed observation model can remarkably discriminate edges of the same vessel than those of different vessels.   

\subsubsection{Generating Probability Maps For the Observation Model \label{sec:GenProbMap}}
 
In this study, we aim to generate probability maps of vessel interior, centerline and edge pixels with a single network. The latter two probability maps are directly used in the likelihood calculation in (\ref{eqn:LikelihoodModel1}) while the former one is utilised to calculate eigenvectors to estimate vessel directions. The network selected was a specific version of Deep Belief Nets (DBNs) \cite{fasel2010deep}, trained to transform one image to another. With this approach, for example, fundus images can be converted to their vessel probability maps, a way of describing the segmentation task. This network was observed to require considerably less training time but producing comparable segmentation performance on the detection of vessel interior pixels in pilot experiments (a performance comparison can be found in Section \ref{sec:Segmentation}), when compared to alternative approaches to the same task (e.g. U-Net \cite{ronneberger2015u} and other CNN based segmentation methods \cite{liskowski2016segmenting,cheng2014discriminative}). 
  
\begin{figure}[!ht]
\centering
{\includegraphics[scale=0.4]{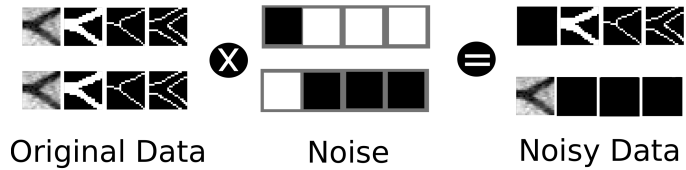} } 
\caption{Generating noisy training samples: A training sample consists of a fundus image and its vessel interior, centerline and edge label maps. In order to generate noisy training samples, a sample can be multiplied with $\left\{0,1,1,1\right\}$ or $\left\{1,0,0,0\right\}$, where $1$ is represented with white squares while $0$ with black squares. \label{fig:CRBM_activation1}}
\end{figure}

\paragraph{Training the network} DBNs are often initially trained as a deep feature extractor in a generative manner, then fine-tuned with stochastic gradient descent to accomplish a target task, usually classification. In this study, a DBN is initially trained in a generative way to learn the relation between fundus images and their label maps (e.g. vessel interior, centerline and edges), which may be expressed as $f: \left\{I_f,  I_i,I_c,I_e\right\}\rightarrow \left\{\overline{I_f,  I_i,I_c,I_e}\right\}$, where $f(\cdot)$ is a function to learn the representation of its input. $\left\{I_f,  I_i,I_c,I_e\right\}$ represents the concatenation of fundus images, $I_f$, vessel interior, $I_i$, centerline, $I_c$ and edge, $I_e$ label maps, respectively, and $\left\{\overline{\cdots}\right\}$ denotes the joint representation learned by the DBN. The concatenation of these images is demonstrated in Figure \ref{fig:CRBM_activation1}, where the concatenated images correspond to a sample in the training dataset. Then, the DBN is fine-tuned to transform fundus images to their label maps by learning a function $g:I_f \rightarrow \left\{ I_i,I_c,I_e\right\}$, where $g(\cdot)$ converts a fundus image to a trio of probability maps representing vessel interior, $I_i$, centerline $I_c$ and edge $I_e$.

\begin{figure} [!ht]%
\centering
{\includegraphics[scale=0.31]{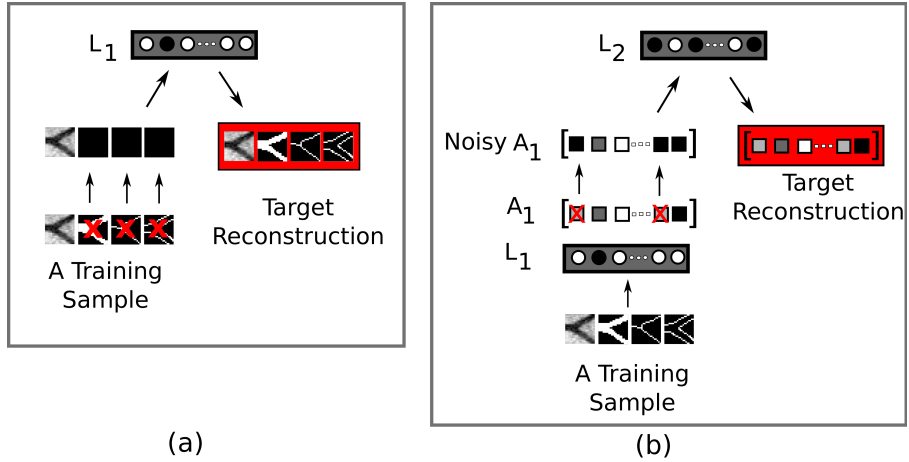}}
\caption{Integrating denoising with the training of a DBN: (a) The training of the first hidden layer $L_{1}$ of the DBN, where denoising is applied image-wise. Red crosses show images to replace with zero for denoising; (b) The training of the second hidden layer $L_{2}$ of the DBN, where denoising is applied unit-wise. Squares illustrate the activations of units in the first hidden layer when they are fed with a training sample. The vector of activations is denoted by $A_{1}$. The units whose activations are suppressed for denoising are shown with red crosses. This type of denoising can be applied to following layers of the DBN (e.g. the third hidden layer). (Figure best viewed in color.)  \label{fig:Denoising}}
\end{figure}

\paragraph{Denoising} 
The interaction between pixels at the same locations, but belonging to either the fundus image or one of its label maps, can be more efficiently learned by combining the generative training of DBNs with the denoising proposed by Vincent \textit{et al.} \cite{vincent2010stacked}.  The spirit of the denoising is to hide some information in the training data on which a network is trained, encouraging the network to predict the missing information in the training data. Sticking to this spirit, we randomly replace either a fundus patch or its label maps with zeros in a training sample, but motivate the network to estimate unaltered pixel values in the sample; $f\big(\left\{0,  I_i,I_c,I_e\right\} \textrm{or} \left\{I_f,  0\right\}\big)=\left\{\overline{I_f,  I_i,I_c,I_e}\right\}$. Because this type of denoising is applied at pixel level, we call it image-wise denoising. Figure \ref{fig:Denoising}(a) shows how to combine the image-wise denoising with the training of the first hidden layer of a DBN. The second and upper hidden layers of the DBN can be trained by applying denoising in a unit-wise manner as originally proposed \cite{vincent2010stacked}, as demonstrated in Figure \ref{fig:Denoising}(b). Denoising is only introduced during generative training. 

\begin{figure} [!ht]%
\centering
{\includegraphics[scale=0.4]{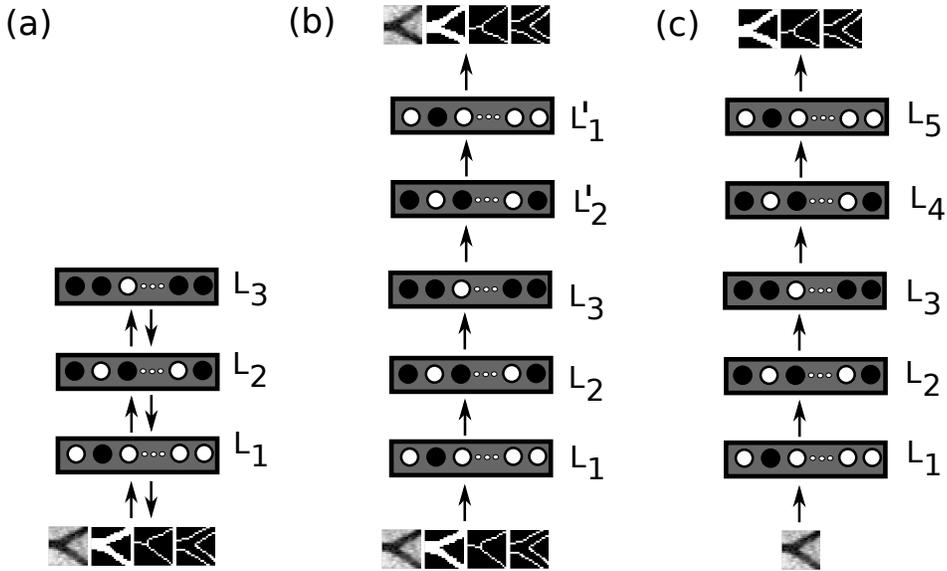}}
\caption{(a) The DBN trained in a generative way as demonstrated in Figure \ref{fig:Denoising}, (b) Unfolding the DBN in (a), (c) Modifying the unfolded DBN in (b) for fine-tuning.  \label{fig:Finetuning}}
\end{figure}

After completing the training, the DBN layer-wise (see Figure \ref{fig:Finetuning}(a)), its weights connecting layers are "unfolded'' \cite{hinton2006}, resulting in a deep autoencoder (see Figure \ref{fig:Finetuning}(b)). Finally, the unfolded DBN is modified by removing weights not contributing to the image transformation task (see Figure \ref{fig:Finetuning}(c)) and, the modified network is fine-tuned with a simple stochastic gradient descent algorithm with $L_{2}$ loss. 

\section{Experimental Setup and Results}
\subsection{Dataset}
The REVIEW dataset \cite{al2008review} contains images collected in the diabetic retinopathy clinic at Sunderland Eye Infirmary during clinical routine.  This dataset has $4$ sub-datasets: the high resolution image set (HRIS), the vascular disease image set (VDIS), the central light reflex image set (CLRIS) and the kick point image set (KPIS). Some properties of the sub-datasets are summarized in Table \ref{table:ReviewSubsets}.

\begin{table}[ht] 
\centering
\caption{The properties of the sub-datasets in the REVIEW}
\label{table:ReviewSubsets}
\resizebox{1\textwidth}{!}{%
\begin{tabular}{ @{}  l c c c c c cc@{}} \toprule
Name &Camera& FOV&Resolution & \multicolumn{2}{c}{No. Of Images}&No. Of& No. Of\\ 
&&&(pixels)&Healthy&Diseased&Ves. Segment&Profiles\\
\cmidrule(lr){1-1}  \cmidrule(lr){2-2} \cmidrule(lr){3-3} \cmidrule(lr){4-4} \cmidrule(lr){5-5}\cmidrule(lr){6-6}\cmidrule(lr){7-7}\cmidrule(lr){8-8}
 HRIS&  Cannon 60 UV& $60^{o}$ &3584 x 2438 & -&4& 90&2368 \\
  VDIS& Zeiss $\&$ JVC 3CCD   & $50^{o}$ &1360 x 1024 &2&6&79&2249\\
  CLRIS& \begin{tabular}{@{}c@{}} Zeiss FF 450 \\ $\&$ JVC 3CCD \end{tabular} & $50^{o}$ &2160 x 1440 &-&2&21&285\\
  KPIS& Canon 60UV& $60^{o}$&3300 x 2600 &2&-&3&164\\
\bottomrule						
\end{tabular}
}
\end{table}

HRIS and KPIS provide a performance evaluation at a sub-pixel accuracy because images in these datasets were down-sampled by a factor of $4$ after receiving estimates of widths by observers; therefore, the accuracy of width estimation was limited to an error of $\pm0.25$ pixels. In HRIS, two of images were graded with severe and one with moderate and the other one with minimal non-proliferative retinopathy. Apart from other sub-datasets, vessel profiles in KPIS were marked through detecting kick points on thicker and non-tortuous vessel segments between bifurcation locations. 

VDIS contains images with pathologies and higher noise. The images were observed to have a larger variance of vessel profiles provided by observers \cite{al2008review}. 6 of the images were captured from patients with various types of Diabetic Retinopathy. CLRIS consists of images showing signs of atherosclerosis, which are the exaggeration of the central light reflex and changes in vessel walls.

\paragraph{Marking Vessel Edges}

Vessel profiles were detected by three experts, two of them with an experience in retinal vessel analysis and the other one trained to locate vessel edges. The experts independently located vessel edges at the same vessel segments. Then, the edge locations were edited by an algorithm to ensure even spaces between neighbor profiles \cite{al2008review}. The average of edge locations marked by the experts is used as reference data to reduce inter-subject variability on detected vessel boundaries.

\subsection{Experimental Settings}
\subsubsection{Overview}
Although the performance of the tracker was evaluated on the REVIEW dataset, this dataset does not include the vessel maps required for segmentation. In order to deal with the problem, we used knowledge transfer, where we trained the network with a well-known fundus image dataset, the DRIVE \cite{staal2004ridge} then used the trained network to generate label maps for the REVIEW. The main challenge with this approach is that the resolution of the REVIEW is much larger than that of the DRIVE. Therefore, the resolution of the REVIEW was reduced by down-sampling before being fed into the network and its resolution was brought the original level prior to being used for tracking. This solution was acceptable for the present research, because the probability maps were only used in the likelihood calculation, which contributes to the calculation of the posterior probability distribution with relative fitness of the hypothesized geometry parameters.  The sub-sampling factors were $2$ for VDIS, $3$ for CLRIS and $4$ for HRIS. No sub-sampling was applied to KPIS.

The centerline and edge images required for training were generated by applying a standard thinning algorithm \cite{lam1992thinning} and a Prewitt edge detection algorithm \cite{shrivakshan2012comparison} respectively to the reference vessel maps. The training of the network was realized patch-wise, where we randomly selected patches from each fundus image map and its corresponding vessel interior, centerline and edge label maps, at the same locations. The size of an image patch was $16$ by $16$ pixels. Because the DRIVE has two vessel maps for each image in its training set, we used the maps produced by the first expert as reference, complying to the general practice \cite{staal2004ridge}.

In order to increase the representation of vessel pixels in the training dataset, we performed denser patch sampling inside the Field-of-View (FOV) masks as follows:  initially, we multiplied FOV masks with the green channels of the fundus images, so pixels outside FOV regions became zero. After randomly and densely sampling image patches, we removed patches completely outside the FOV masks. The number of the patches in the final stage was roughly $1,800,000$. The fundus image patches were normalized patch-wise in the range of $[0,1]$, which was visually observed to better reveal vessels on patches with lower contrast. 

\subsubsection{Network Parameters}
Because the network goes through different types of training, the configuration of the network is altered accordingly; this is typical of DBN methods \cite{hinton2006reducing,fasel2010deep}. Initially, the network has the input layer of $256 \times 4=1024$ units and $3$ hidden layers, with each having $400$ units during the generative training. After 'unfolding' and reshaping it for fine-tuning, the network contains the input layer of $256$ units, the output layer of $256 \times 3=768$, units and $5$ hidden layers of $400$ units. 

The network was trained with mini-batches of $100$ sets of image patches for both the generative training and fine-tuning. In the generative training, the network weights were initialized by sampling from a normal distribution $\mathcal{N}(0,0.001)$. Then, the network was trained for $50$ epochs with a learning rate of $0.005$. A momentum of $0.5$ was initially used in the first $5$ epochs. Later, this number was increased to $0.9$. The learning rate for fine-tuning was $0.08$ for $120$ epochs. The squeezing function for all layers was sigmoid.  

\subsubsection{Tracking Parameters}
The number of particles for tracking was $700$ and step size was $2$ pixel. Vessel centerline and width parameters or each vessel segment were initialized with the reference data in the REVIEW. In the same way, vessel direction is assigned the direction from the first reference centerline location to second one, similar to \cite{zhang2014retinal}.

\subsection{Evaluation Criteria \label{sec:EvaluationCriteria}}
\textit{Precision}, in  (\ref{eqn:Precision}), and \textit{accuracy}, in (\ref{eqn:Accuracy}), are the most commonly used measures to evaluate the fitness of estimates of vessel widths to reference widths \cite{araujoa2017estimation,aliahmad2016adaptive,xu2011vessel, al2009active}. 
\begin{align}
&Precision = \textrm{std}(w_{r}-w_{e}) \label{eqn:Precision} \\
&Accuracy= \textrm{mean}(|w_{r}-w_{e}|)  \label{eqn:Accuracy} 
\end{align}
where $w_{r}$ denotes reference widths while $w_{e}$ shows their estimates.

In vessel width estimation, relative widths, rather than actual widths, over a fundus image are usually used in assessment \cite{lowell2004measurement}. One good argument  for this is to have relative assessments of vessel diameter changes (within subject) that are approximately independent of optical magnification. Al-Diri {\it et al.} \cite{al2008review} also points out that consistent biases or scale factors in measurement can be removed by simple linear transforms of width estimates. In line with other studies, we use \textit{Precision} to assess the success of a method, rather than \textit{Accuracy} because of the possible effect of constant bias on the \textit{Accuracy} measure. Therefore, even though \textit{Accuracy} is reported in following experiments, it is to inform readers, not to compare the performance of the methods. 

In addition to \textit{Precision} and \textit{Accuracy}, the percentage of vessel profiles whose width estimates are meaningful was also reported in previous studies \cite{lowell2004measurement,lupacscu2013accurate,aliahmad2016adaptive,xu2011vessel,al2009active,zhou1994detection}. This measure indicates the ability of a method to deal with different challenges (e.g. noise, pathology) without compromising the performance of width estimation \cite{al2009active}. 
\section{Results}

\subsection{Generated Probability Maps by the Network} 
\begin{figure}[!ht]
\centering
    {\includegraphics[width= 11 cm, height= 5.5 cm]{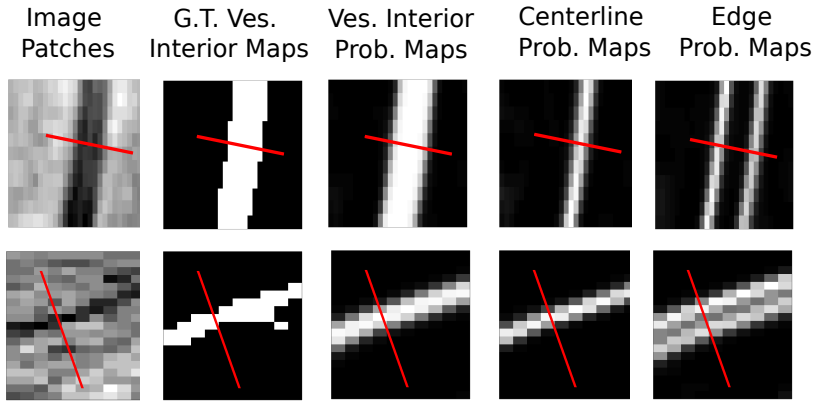} }
   \caption{Fundus image patches and the typical probability maps generated by the deep network: the top row belongs to a large vessel and bottom row to a fine vessel. Columns from the left to right side show a normalized fundus image patch, its manually labeled vessel interior map, generated vessel interior, centerline and edge probability maps. Red lines show the locations that profiles in Figure \ref{fig:VesselProfileComparision} are obtained.  (Best viewed in color.) \label{fig:LabelPatches} }
\end{figure}

The trained network is used to produce probability maps for vessel interior, centerline and boundaries as shown in Figure \ref{fig:LabelPatches}, where two image patches, each containing a thick or a thin vessel, are shown, along with their aforementioned probability maps. Because the patches are normalized in the range of $[0,1]$, the contrast of vessels seems very similar despite the significant difference between their thickness. Thus, despite varying vessel thickness and noise levels, the network-generated smooth probability maps are consistent with the ground truth vessel masks.

\subsection{Segmentation \label{sec:Segmentation}}

Because there is no direct way to asses how well probability maps represent vessel parts in an image, we consider the performance of the network on vessel interior segmentation as an overall indicator of its performance on probability maps. Table \ref{table:SDBNdrive} compares the performance of the network and that of recent state of the art methods using supervised methods, regarding AUC, accuracy, sensitivity and specificity, whose definitions can be found in \cite{Li2015}. Referring to the table, the present study is among the best performing studies regarding AUC.  Apart from other measures, namely accuracy, sensitivity and specificity, AUC is calculated {\em independently} of a subjective threshold, and shows the certainty of the method on discriminating vessel pixels from non-vessel pixels; in other words, it is an indication of the robustness and quality of the vessel interior probability maps.

\begin{table}[ht] 
\centering
\caption{The comparison of segmentation performance of the proposed method with the performances of state of the art methods regarding the DRIVE dataset}
\label{table:SDBNdrive}
\resizebox{0.85\textwidth}{!}{%
 \begin{threeparttable}
\begin{tabular}{ @{}  l l c c c c @{}} \toprule
Year & Method	&AUC	&Accuracy &Sensitivity &Specificity \\ \cmidrule(lr){1-1}  \cmidrule(lr){2-2} \cmidrule(lr){3-3} \cmidrule(lr){4-4} \cmidrule(lr){5-5} \cmidrule(lr){6-6} 
 2017& The proposed method & {0.9761} &    0.9542 &   0.7752 &    0.9800  \\ 
  2017&U-net \cite{ronneberger2015u} \tnote{*} &\textbf{0.9790 }&-&-&-\\
 2016 &Liskowski and Krawiec \cite{liskowski2016segmenting}  & 0.9710  &  0.9515     &0.7520     & 0.9806 \\ 
 2015&Wang \textit{et al.} \cite{wang2015hierarchical} & 0.9475& \textbf{0.9767}& \textbf{0.8173} & 0.9733\\ 
 2015 &Li \textit{et al.}\cite{Li2015} 	& 0.9738	& 0.9527	& 0.7569	& \textbf{0.9816}\\ 
 2014 &Cheng \textit{et al.}  \cite{cheng2014discriminative} & 0.9648	& 0.9474	& 0.7252	& 0.9798	\\ 
  2012 &Fraz \textit{et al.} \cite{fraz2012ensemble}	& {0.9747}	&0.9480	&0.7406	&0.9807		\\ 
\bottomrule						
\end{tabular}
\begin{tablenotes}
  \item[*] The result was taken from \url{https://github.com/orobix/retina-unet} not from \cite{ronneberger2015u}.
 \end{tablenotes}
\end{threeparttable}
}
\end{table}

\subsection{Width Estimation}
We now assess the performance of the tracker on the estimation of vessel width, in the REVIEW dataset. The conventional way to evaluate the reliability of vessel width estimation is to compare estimated widths to the reference ones over predetermined profiles where the estimated and reference profiles either share only the same centerline locations or both centerlines and orientations \cite{lupacscu2013accurate,araujoa2017estimation,aliahmad2016adaptive,xu2011vessel,al2009active}. However, such a one to one comparison was not possible for the current method because both centerline locations and the orientations of the vessel cross-sections were autonomously estimated by the tracker.

In order to reduce a potential discrepancy between the locations of widths estimated by the method and those given in the reference data as much as possible, we used  bi-cubic spline interpolation to sample $100$ locations, from both reference and estimated widths. Then, the evaluation criteria were calculated over interpolated values. However, it should be noted that because the locations of interpolated profiles from both reference and estimations were not guaranteed to overlap, this evaluation also has potential limitations. Also, vessel segments with less than $2$ reference profiles could not be traced because these profiles are either used to start tracking or to stop it.

\subsubsection{Evaluation Over A Dataset}
Traditionally, the performance of width estimation is compared over all profiles in the sub-dataset, regardless of vessel identification \cite{araujoa2017estimation,aliahmad2016adaptive,xu2011vessel, al2009active}. Table \ref{table:ReviewdatasetWidths} compares the performance of the proposed method to that of previous studies regarding \textit{Accuracy}, \textit{Precision} and the percentage of meaningful width estimations. The proposed method is able to obtain meaningful estimates of width for the entire set of vessel profiles, whereas the majority of methods fails to predict plausible widths for some profiles in CLRIS, HRIS and VDIS; this includes the best performing methods of Zhang \textit{et al.} \cite{zhang2014retinal} and Yin \textit{et al.} \cite{yin2012retinal}.  

Regarding \textit{Precision}, the most successful results were reported by Ara{\'u}joa \textit{et al.} \cite{araujoa2017estimation}, who used a supervised model-fitting method. However, because they did not present the percentage of meaningful width estimations, it is not clear if the results reflect the \textit{Precision} of overall sub-datasets, or only those corresponding to the successfully estimated profiles. The performance of Ara{\'u}joa \textit{et al.}'s method \cite{araujoa2017estimation} is followed by that of tracking methods \cite{yin2012retinal,zhang2014retinal}, whose performance mostly surpasses that of other supervised and unsupervised methods \cite{lowell2004measurement,lupacscu2013accurate,xu2011vessel,al2009active,zhou1994detection}.

Amongst tracking approaches (see Table~\ref{table:ReviewdatasetWidths}), the performance of the proposed method closely follows that of Zhang \textit{et al.}'s method regarding \textit{Precision}. This dataset has been generally found challenging by many methods \cite{zhang2014retinal}, due to containing the central light reflex along vessel segments. Regarding HRIS, \textit{Precision} of the proposed method is slightly worse than that of Zhang \textit{et al.} \cite{zhang2014retinal} and that of Yin \textit{et al.} \cite{yin2012retinal}. For VDIS, the proposed method has larger \textit{Precision} than that of Zhang \textit{et al.} \cite{zhang2014retinal} and that of Yin \textit{et al.} \cite{yin2012retinal} but the proposed method predicts all widths for this dataset while Zhang \textit{et al.} \cite{zhang2014retinal} and Yin \textit{et al.} successfully estimated only $94.2\%$ and $92.7\%$ of widths respectively. For KPIS, the proposed method gives the lowest \textit{Precision} among tracking methods, which is in the same range as that among supervised and unsupervised methods.

\begin{table}[ht] 
\centering
\caption{The performance comparison of the proposed method with those of previous studies on the REVIEW dataset, where $\%$ represents the percentage of meaningful width estimations}
\label{table:ReviewdatasetWidths} 
 \begin{threeparttable}
 \resizebox{1\textwidth}{!}{%
\begin{tabular} {@{} llcccc ccccccccc@{} } \toprule %
&& &
      \multicolumn{3}{c}{CLRIS} &
      \multicolumn{3}{c}{HRIS} &
      \multicolumn{3}{c}{VDIS}  &
      \multicolumn{3}{c}{KPIS}\\  
      \cmidrule(lr){4-6}  \cmidrule(lr){7-9}  \cmidrule(lr){10-12}  \cmidrule(lr){13-15}
&Method         &Year  & Acc. & Prec.  & \%& Acc. & Prec.&  \%& Acc. & Prec.&  \%& Acc. & Prec.  &  \%     \\  \cmidrule(lr){2-2} \cmidrule(lr){3-3} \cmidrule(lr){4-4} \cmidrule(lr){5-5} \cmidrule(lr){6-6} \cmidrule(lr){7-7} \cmidrule(lr){8-8} \cmidrule(lr){9-9} \cmidrule(lr){10-10}  \cmidrule(lr){11-11}  \cmidrule(lr){12-12}   \cmidrule(lr){13-13} \cmidrule(lr){14-14}\cmidrule(lr){15-15}
\cmidrule(lr){14-14}              
&Observer $1$     &  &   0.61   &0.57  & &  0.23& 0.29  &&0.35& 0.54  && 0.34 & 0.42 &         \\
&Observer $2$      & &   0.11 & 0.70 && 0&0.26 && 0.06 & 0.62  && 0.11 & 0.32   &       \\
&Observer $3$       &&   0.72 & 0.57 &&0.23 & 0.29 &&0.3&0.67 && 0.23 &0.33 &  \\ \midrule
\multirow{ 5}{*}{Tracking M.} 
&The proposed method &2017& 0.92  &  1.15 &\textbf{100}&0.32 &    0.40&\textbf{100}&1.01  &  0.85&\textbf{100}&1.51 &   \textbf{0.34}&\textbf{100}\\ 
&Zhang \textit{et al.} \cite{zhang2014retinal} & 2014& 0.37&\textbf{1.13}  &98.3 &0.08&\textbf{0.30}   & \textbf{100}&1.37&{0.59} &94.2& 0.74& {0.37}&   \textbf{100} \\
&Yin \textit{et al.} * \cite{yin2012retinal} &2012&0.77 & 1.41  & 93.1 &0.01 &0.39 &96.2&1.41 & \textbf{0.56 }&92.7&  0.69&0.43& \textbf{100 }   \\ 
& Zhou \textit{et al.} ** \cite{zhou1994detection} &1994&7.5  & 4.14 &98.6&0.54 & 0.90 &99.6&3.07 &2.11&99.9&   2.57 &0.4&  \textbf{ 100} \\   \midrule
\multirow{ 5}{*}{Supervised M.}&Ara{\'u}joa \textit{et al} \cite{araujoa2017estimation} & 2017& {0.01}&\textbf{0.56}&-  &{0.00}&\textbf{0.22}&- &  {0.00}&\textbf{0.69} &-& {0.00} &\textbf{0.30} & -  \\
&Aliahmad and Kumar \cite{aliahmad2016adaptive} & 2016& 0.33 &1.56  &98 &0.24 &0.65 &99.4&0.45 &1.14&97.8& 0.72&0.45 &\textbf{100}   \\
&Lupa{\c{s}}cu \textit{et al.} \cite{lupacscu2013accurate} & 2013 & 0.00& 1.15&\textbf{100}& 0.00 & 0.44&\textbf{100}&0.02 & 1.07 & \textbf{100}&0.02 & 0.32 & \textbf{100}\\ \midrule
\multirow{ 5}{*}{Unsupervised M.} &Xu \textit{et al.}    \cite{xu2011vessel} &2011& 0.08 & 1.78 &94.3& 0.21 &0.567 &\textbf{100}& 0.53 &1.43&96&1.14 &0.67 &99.4    \\
&Al-Diri  \textit{et al.}  \cite{al2009active} & 2009&1.9 & \textbf{1.47} & 93 &0.28 &\textbf{0.42}& 99.7& 0.05&\textbf{0.77} &99.6& 0.96&\textbf{0.33}& \textbf{100}  \\
&Lowell \textit{et al.} **   \cite{lowell2004measurement} & 2004& 6.8& 6.02 &26.7&0.17 &0.70&98.9&  2.26&1.33&77.2&1.65&0.34&  \textbf{100} \\
 
\bottomrule
\end{tabular}
}
\begin{tablenotes}
 \item[*] These results were taken from \cite{zhang2014retinal}.
 \item[**] These results were taken from \cite{aliahmad2016adaptive}.  
 \end{tablenotes}
 \end{threeparttable}
\end{table}

\subsubsection{Evaluation For Each Vessel Segment}

In contrast to the traditional performance evaluation, which implicitly accepts that vessel profiles are independently sampled from a dataset and summarizes the performance with a single number, we also assessed the performance of the proposed approach for each vessel segment in REVIEW, based on the fact that profiles selected from the same vessel segment are highly probable to have similar widths, and also similar types of problems, such as the presence of the central light reflex. Therefore, we calculated \textit{Precision} for each vessel segment and obtained \textit{Precision} distribution for each sub-dataset to observe if poor or good performance of the method on the sub-dataset may be related to the performance on specific vessel segments. 

\begin{figure}[!ht]
\centering
  \begin{subfigure}[b]{1\textwidth}
  \centering
 {\includegraphics[scale=0.3]{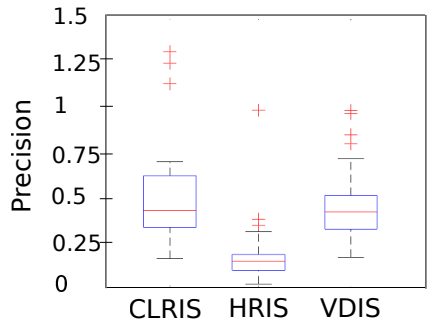}}
 \end{subfigure}
  \caption{The distributions of the \textit{Precisions} of vessel widths calculated for individual vessel segments in CLRIS, HRIS and VDIS. The maximum length of each whisker is $1.5$ times of the interquartile range of related distribution.(Best viewed in color.) \label{fig:PredictionReferencePlotAllAutomatic} }
\end{figure}

Figure \ref{fig:PredictionReferencePlotAllAutomatic} shows the distributions of the \textit{Precisions} of vessel widths  produced for CLRIS, HRIS and VDIS with box-plots. Because KPIS does not have a sufficient number of vessel segments for this demonstration, its results are, instead, summarized in the text. The figure shows three outliers for both CLRIS and HRIS, and four outliers for VDIS,with \textit{Precisions} of over $1$ pixel, which indicates abnormal disagreements between reference and estimated widths on the vessel segments responsible for the outliers. It should be noted that this information was not revealed in Table \ref{table:ReviewdatasetWidths}. The \textit{Precisions} for CLRIS, HRIS and VDIS in the table are, respectively, $1.15$, $0.40$ and $0.85$ pixels, which are far larger than the medians of the \textit{Precision} distributions illustrated in the figure. Also, according to the figure, the median \textit{Precisions} for CLRIS and that for VDIS are almost the same; however, the \textit{Precisions} of outliers in CLRIS are much larger than those in VDIS. Obviously, these extreme outliers can be taken responsible for CLRIS with larger \textit{Precision} than \textit{Precision} of VDIS in Table \ref{table:ReviewdatasetWidths}.

In addition to results presented in the figure, KPIS was observed to have the \textit{Precisions} of $0.35$, $0.31$ for the first and second vessel segment in the first image respectively and, that of $0.37$ for the single segment in the second image. These results are mainly consistent with the \textit{Precision} given in Table \ref{table:ReviewdatasetWidths}.

We argue that the proposed way of calculating the evaluation criteria is more appropriate than the traditional way \cite{araujoa2017estimation,aliahmad2016adaptive,xu2011vessel, al2009active}, because the former can identify vessel segments for which a given method yields significantly different widths from reference data. The identification of the challenging vessel segments in this manner is useful when developing new approaches to tracking or width estimation. We will closely examine vessel segments that have been found to be challenging in order to appreciate the sources of disagreement between our estimations and the reference data.

\begin{figure}[!ht]
\centering
 {\includegraphics[scale=0.25]{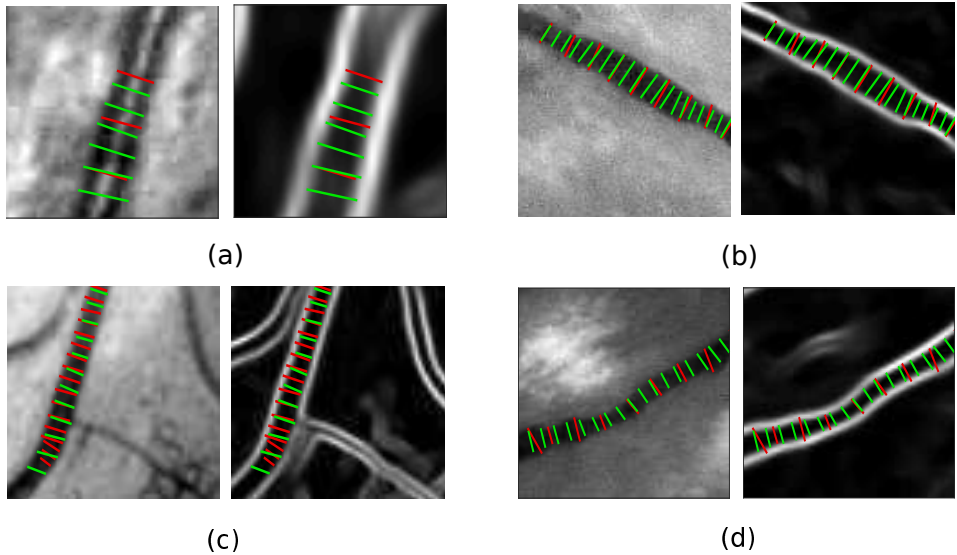}}
  \caption{Interpolated reference, shown with red lines, and estimated vessel profiles, shown with green lines, for vessel segments with the worst performance of the proposed method: (a) segment $2-8$ in CLRIS, (b) segment $8-3$ in VDIS, (c) segment $1-26$ in HRIS, (d) segment $8-10$ in VDIS. The left and right images for each image couple respectively show a fundus image and its edge probability map. The images in the top row have the size of $60 \times 60$ pixels and $122 \times 122$ pixels respectively while those in the bottom row have the size of $66 \times 66$ pixels and $100$ by $100$ pixels consecutively. (Figure best viewed in color.) \label{fig:Outliers1} }
\end{figure}

\paragraph{Outliers in the Box-Plot} The vessel segments producing largest \textit{Precision} for HRIS, VDIS and KPIS in Figure \ref{fig:PredictionReferencePlotAllAutomatic} are demonstrated with estimated widths and reference ones on both fundus images and edge probability maps in Figure \ref{fig:Outliers1}. 

Figure \ref{fig:Outliers1}(a) shows a vessel segment from CLRIS, with the \textit{Precision} of $1.3$. The subtle change on vessel width seems not to be captured by human observers, in contrast to the proposed method. 
This situation can also be observed in Figure \ref{fig:Outliers1}(b), which demonstrates an image from VDIS, with the \textit{Precision} of $0.98$. This vessel segment can be characterized with abnormal width changes. Similar to former image pair, vessel widths are estimated reasonably consistently by the proposed method, following the changes on actual vessel width in pathologies.

The superiority of the current method to human observers becomes more obvious in Figure \ref{fig:Outliers1}(c), where an image pair from HRIS, with the \textit{Precision} of $0.98$, is demonstrated. The vessel edges in this figure are mis-detected by the observers.

Apart from imperfection of human observers to accurately estimate vessel widths, the discrepancy between estimates and reference data can also be due to predicting widths along slightly different profiles as appearing in Figure \ref{fig:Outliers1}(d). This figure illustrates a vessel segment from VDIS, with the \textit{Precision} of $0.97$.

\subsubsection{Challenging Vessels \label{sec:ChallengingVessel}}
In the literature, some vessels have been found more challenging than others; such as (i) those with the central light reflex \cite{lowell2004measurement,al2009active}, (ii) close vessels \cite{al2009active,bekkers2014multi}, (iii) those in junction regions \cite{zhou1994detection,bekkers2014multi} and (iv) those with high curvature or,  (v) low contrast or  high noise. We now discuss each of these in turn.

\begin{figure}[!ht]
  \centering
 {\includegraphics[scale=0.25]{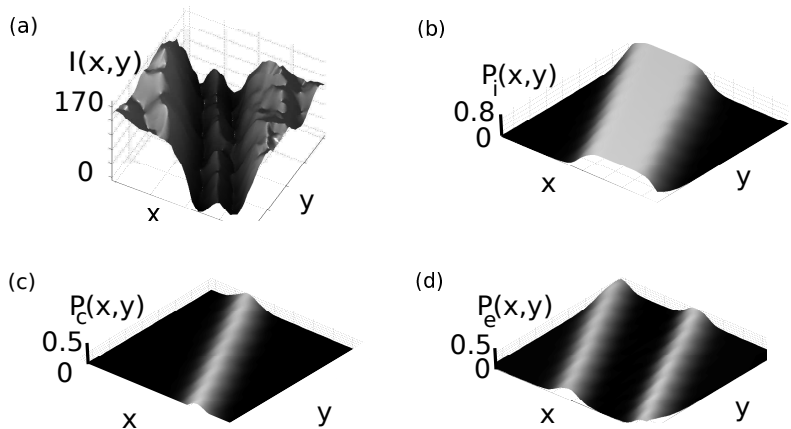}}
  \caption{The intensity profile $I(x,y)$ of a vessel with the central light reflex, a part of the vessel in Figure \ref{fig:Disagreement}(a), and its probability profiles $P_{i}(x,y)$, $P_{c}(x,y)$ and $P_{e}(x,y)$, which respectively denote vessel interior, centerline and edge locations.  \label{fig:3Dprofiles} }
\end{figure}

\begin{figure}[!ht]
  \centering
 {\includegraphics[scale=0.25 ]{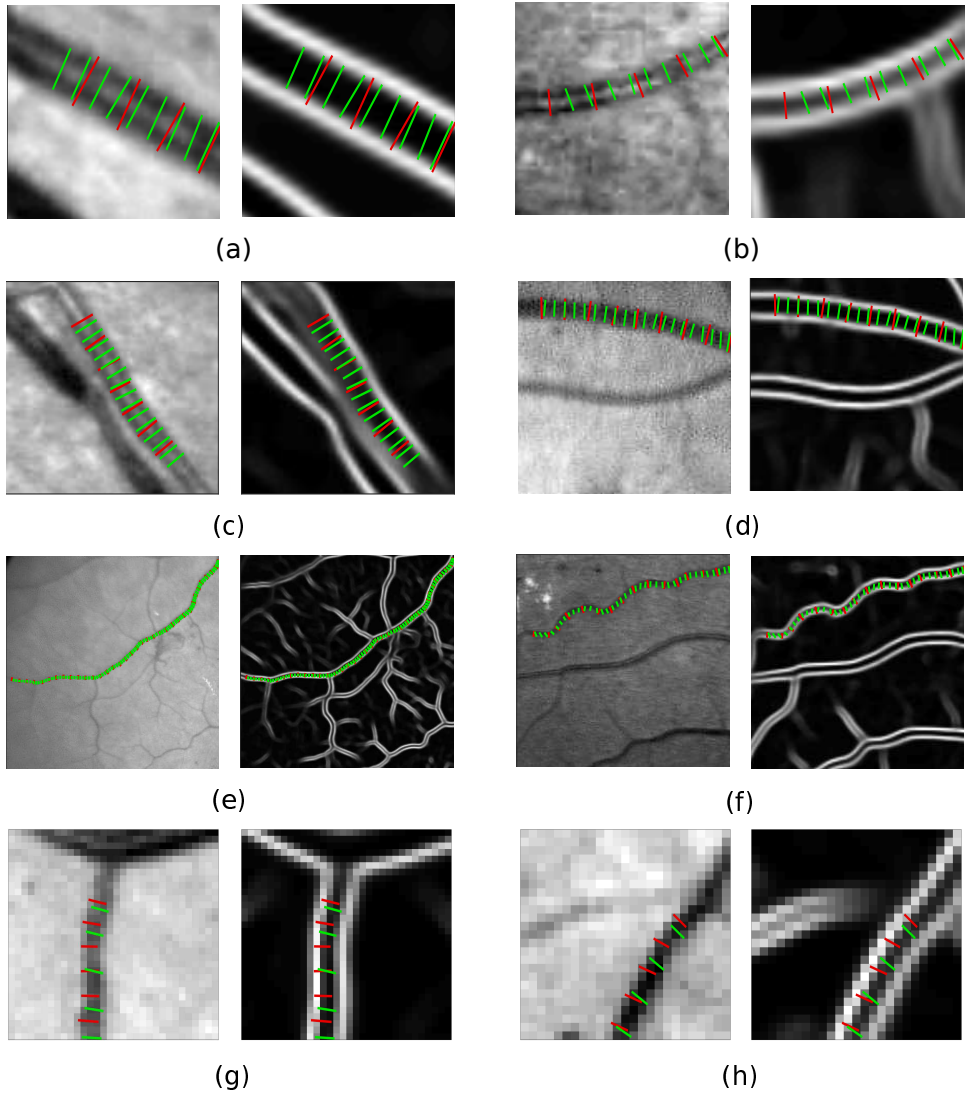}}
  \caption{Interpolated reference, shown in red, and estimated profiles by the proposed method, shown in green, for challenging vessels:  (a)-(c) from CLRIS, (d)-(f) from VDIS and (g)-(h) from HRIS. The left and right images for each image couple respectively show a fundus image and its edge probability map. The sizes of images are respectively $63$ by $63$ pixels in (a), $64$ by $64$ pixels in (b), $98$ by $98$ pixels in (c), $130$ by $130$ pixels in (d), $424$ by $424$ pixels in (e), $207$ by $207$ pixels in (f), $31$ by $31$ pixels in (g) and $25$ by $25$ pixels in (h). (Best viewed in color.) \label{fig:Disagreement}}
\end{figure}

(i) The central light reflex is a bright strip around vessel centerline, which may be confused with vessels edges by methods \cite{chapman2001computer}. In the presence of the reflection, the intensity profile across vessel edges deviates from Gaussian-like appearance, which may be compensated by increasing the complexity in models for the intensity profile, such as representing it with multiple Gaussian functions \cite{lupacscu2013accurate,araujoa2017estimation,gao2001method}. Another way may be to combine a method using the intensity profile as the main information source with additional sources, for example line detector responses \cite{zhang2014retinal}. On the other hand, the proposed method does not need to take extra measures to deal with the reflection, which is naturally suppressed during the network training. Figure \ref{fig:3Dprofiles} shows a vessel with the central light reflex from CLRIS and the probability maps generated by the network, where no sign for the reflection appears. Figure \ref{fig:Disagreement}(a)-(c) show estimated and reference profiles for vessels with the reflection. As may be seen, the presence of the reflection does not degrade the consistency of width estimations.

(ii) In some cases, vessels can be close to each other which makes it difficult to estimate vessel geometry: it  may not possible to identify vessel edges due to the presence of the other nearby vessel \cite{al2009active,bekkers2014multi}. However, the proposed method can be observed to successfully trace a vessel with the central light reflex and low contrast, without being distracted by a closely passing one, as shown in Figure \ref{fig:Disagreement}(c). When the figure is examined closely, it appears that edge probability map of the traced vessel is affected by the nearby vessel to a large extent, which is manifested with far lower and diffused probabilities for the left edge of the traced vessel. Despite the large uncertainty along this edge, the tracker manages to identify both vessel boundaries correctly and tracks the vessel without any disruption. This may be attributed to the existence of the prior probability distribution keeping the memory of the previously traced path. 

(iii) Vessels in junction regions are difficult to analyze because vessel boundaries may become completely indistinguishable  \cite{zhou1994detection,bekkers2014multi}. The edge pixels in junctions can have lower edge probabilities, which is visible in Figure \ref{fig:Disagreement}(e). Because of the prior information implicit in the tracking process, the region of lower edge probabilities can be traced confidently.

(iv) Curvy vessels may pose a big challenge for tracking methods due to their fast changing directions. However, vessels with high curvature are observed to not pose significant problems for the proposed method, because of the use of estimated vessel direction, and incorporated in the $P_s$ term of \ref{eqn:LikelihoodModel1}). Figure \ref{fig:Disagreement}(f) shows estimated widths for a curvy vessel. 

(v) Human observers can also fail at estimating the right locations for vessel edges, particularly, if the contrast of vessels is poor or their calibers are small. Figure \ref{fig:Disagreement}(d),(g)-(h) illustrate profiles located by the observers and the present method. On fundus images, both reference and estimated locations seem acceptable to the naked eye. This may align with the inter-observer variability in locating vessel boundaries on ground truth images, a problem acknowledged in \cite{staal2004ridge,hoover2000locating}. However, considering the disagreement between the values in the edge probability maps at the edge locations estimated by the observers (see Figure~\ref{fig:Disagreement}(g)-(h)) and those of the proposed tracker, we argue that our method estimates better edge locations in these images than human observers.

\section{Conclusion}
In this study, we propose a Bayesian method \footnote{The code producing results reported in this paper can be found at \url{https://bitbucket.org/fzehra/a-recursive-bayesian-approach-to-describe-retinal-vasculature}} to estimate vessel geometry parameters, and evaluated the performance of the method on width estimation on the REVIEW dataset. In contrast to previous methods, which have used the intensity profiles across vessel edges for parameter estimation, we utilized probability profiles sampled from vessel interior, centerline and edge location probability maps generated by a single deep network.  As far as we are aware, it is the first method for retinal vessel analysis -- or any vascular data -- that uses probability maps as inputs to a tracking and vessel width estimation process. The method addressed four challenges encountered in width estimation. 

The first is that probability maps for vessel parameter estimation can better explain the uncertainty and subjectivity at detecting vessels and, particularly, edge locations, appearing in the ground truth data \cite{staal2004ridge,al2008review}. Due to these maps serving a Bayesian method, particle filtering, the uncertainty in these maps is efficiently utilized for vessel parameter estimation. For instance, the proposed approach could make reasonable estimates of vessel width: even when there is not sufficient information available for vessel parameter estimation (e.g. in junction regions), or when the information is vague (e.g. for thin and low contrast vessels, and vessels in a close proximity to another vessel). Estimating all sets of vessel profiles in the REVIEW reinforces the effectiveness of this approach.  Moreover, having edge probability maps facilitated an evaluation of the consistency of reference vessel profiles.

Secondly, despite the lack of training data for vessel segmentation specific to the REVIEW dataset, the proposed method was able to generate useful probability maps for vessel geometry. Specifically, training the network with low resolution and almost healthy fundus images, provided in the DRIVE, was observed to produce sufficient quality of probability maps for the REVIEW, which has high resolution and mostly pathological images. This success may be attributed to two factors. Firstly, the generalization capability of the network was adequate for the purpose of the presented method. Secondly, using particle filtering for vessel parameter estimation might compensate any imperfections in the probability maps. 

Thirdly, to date, the performance of a method on width estimation has been assessed with the independent evaluation of the profiles. However, this approach ignores the spatial dependence of vessel profiles for a particular vessel segment.  In this study, we also assessed the performance of the method on individual vessel segments, which allowed us to immediately spot  disagreements between reference measurements in the REVIEW data and estimates from the proposed method at the level of vessel segments. Moreover, we could evaluate the reliability of the reference data to some extent: our analysis seems to have uncovered some errors in the reference data, revealed by our way of performance evaluation. 

Finally, due to it being independent of the reference data, the proposed method can be viewed as being superior to methods based on supervised model fitting \cite{lupacscu2013accurate,araujoa2017estimation,aliahmad2016adaptive}, whose performance strongly depends on the characteristics of the training datasets. If the datasets have any errors or bias in their reference data, the estimates made by the supervised methods \cite{lupacscu2013accurate,araujoa2017estimation} are highly probable to have the same issues, despite having close agreement with reference data. 

Currently, this work only considers tracing vessel segments for parameter estimation. However, we are working on a method to detect junction locations in fundus images, which will be integrated with the present method to trace complete vessel trees.  
 
 \section{Acknowledgement}
The authors thank to  Republic of Turkey Ministry Of National Education for their financial support.

\bibliographystyle{model1-num-names.bst}
\bibliography{main}

\end{document}